\documentclass{article}

\usepackage{arxiv}
\usepackage[utf8]{inputenc} % allow utf-8 input
\usepackage[T1]{fontenc}    % use 8-bit T1 fonts
\usepackage{hyperref}       % hyperlinks
\usepackage{url}            % simple URL typesetting
\usepackage{booktabs}       % professional-quality tables
\usepackage{amsfonts}       % blackboard math symbols
\usepackage{nicefrac}       % compact symbols for 1/2, etc.
\usepackage{microtype}      % microtypography
\usepackage{graphicx}
\usepackage{gensymb}        %\degree
\usepackage{amsmath}        %\eqref
\usepackage{algorithm}
\usepackage[noend]{algpseudocode}
\usepackage{diagbox}
\usepackage{transparent}
\newsavebox{\strel}

\makeatletter
\newcommand{\shorteq}{%
    \settowidth{\@tempdima}{--}% Width of hyphen
    \resizebox{\@tempdima}{\height}{=}%
}
\makeatother

\title{Complexity of Shapes Embedded in ${\mathbb Z^n}$ with a Bias Towards Squares}

\author{
    Mazlum~Ferhat~Arslan\\
    Department of Computer Engineering\\
    Middle East Technical University\\
    06800 Ankara, Turkey\\
    \texttt{ferhata@metu.edu.tr}\\
    \And
    Sibel~Tari\\
    Department of Computer Engineering\\
    Middle East Technical University\\
    06800 Ankara, Turkey\\
    \texttt{stari@metu.edu.tr}
}

\begin{document}
\maketitle

\begin{abstract}
    Shape complexity is a hard-to-quantify quality, mainly due to its relative nature. 
    Biased by Euclidean thinking, circles are commonly considered as the simplest.
    However, their constructions as digital images are only approximations to the ideal
        form.
    Consequently, complexity orders computed in reference to circle are unstable. 
    Unlike circles which lose their circleness in digital images, squares retain their
        qualities.
    Hence, we consider squares (hypercubes in $\mathbb Z^n$) to be the simplest shapes
        relative to which complexity orders are constructed.
    Using the connection between $L^\infty$ norm and squares we effectively encode
        squareness-adapted simplification through which we obtain multi-scale complexity
        measure, where scale determines the level of interest to the boundary.
    The emergent scale above which the effect of a boundary feature (appendage) disappears
        is related to the ratio of the contacting width of the appendage to that of the
        main body.
    We discuss what zero complexity implies in terms of information repetition and
        constructibility and what kind of shapes in addition to squares have zero
        complexity.
\end{abstract}

% keywords can be removed

\section{Introduction}
\label{sec:intro}
\noindent
Quantifying shape complexity is a classical yet unsolved problem. The key source of
    difficulty is that complexity is a relative concept.
In general, for a given object, the complexity has to do with how it is of interest and
    what tools are available to describe it ({\sl i.e.} it is easier to construct a circle
    if compasses are available, or a triangle if rulers are available).

In the vast majority of the works in the literature, it is assumed that circles are the
    simplest shapes.
The geometric definition and properties of circle bring forth a strong motivation for it
    to be considered as the simplest.
In Euclidean space, one number (radius) is sufficient to construct a circle. 
However, the property that each point of the circle being equidistant to an enclosed
    center as measured under the Euclidean metric is not satisfiable in digital spaces.
Hence, identifying geometric notions of circularity such as uniform curvature or minimal
    perimeter per area do not extend when circles are represented as digital images
    \cite{Rosenfeld1974}, {\sl i.e.}, as subsets of $\mathbb Z^2$.
In this sense, circles lose their {\sl circle-ness}.
Consequently, any measurement obtained by taking circle or circularity as reference is
    inherently unstable.
    
In ${\mathbb Z^2}$, unlike circles which lose their {\sl circle-ness}, rectangles with
    sides parallel to grid axes do not lose their {\sl rectangle-ness}.
An approximation of such a rectangle is still a rectangle; though, it may not be the same
    one because the aspect ratio of the approximated rectangle may be distorted.
Nonetheless, {\sl rectangle-ness} is unharmed.
Of specific interest among rectangles is the one with unit aspect ratio namely square, for
    it also preserves its unit aspect ratio under approximations.
While circles satisfy the property of being equidistant to an enclosed center as measured
    under Euclidean metric, squares satisfy the same property under $\| . \|_\infty$
    metric.
The arguments can be extended to arbitrary dimensional shapes. 
Therefore, we formulate our shape complexity measure with the bias that hypersquares are
    the simplest shapes by choosing $\left( \mathbb{Z}^n,\,\| . \|_\infty \right)$ as the
    right space to measure interactions among a digital shape's atomic elements, pixels.

%$\left( \mathbb{P} (\mathbb{Z}^n),\,\| . \|_\infty \right)$ 
Any shape complexity measure should be multi-scale.
This is important because a given shape data may contain noise or a shape may have
    features that are only meaningful at certain scales.
Hence, a key ingredient of any shape complexity measurement method is a coarsening
    process.
For the consistency and robustness of the complexity measure, the coarsening process
    should be suitable with the chosen metric space, meaning that shapes are expected to
    flow towards the simplest shapes that can be constructed in the chosen space.
This is the main drive behind our constructions which use $L^\infty$ space to construct
    functions that represent gradual coarsening of the shape boundary.
In our multi-scale framework, the scale interval $[0,1]$ can be sampled up to a number of
    discrete scale intervals not exceeding the {\sl shape width} as measured in 
    $\| . \|_\infty$.
At coarser scales, the sensitivity to boundary details is lost causing different
    shapes (or shapes with different noise levels) to converge in complexity. 
Indeed, in our method, a cutoff scale emerges such that above it the effect of an
    appendage (boundary noise or detail) disappears.
For rectangular compositions, this cutoff scale is related to the ratio of the
    contacting width of the appendage to the width of the main body
    (\S~\ref{ssec:congruence}).
Furthermore, when integrated over \mbox{low-,} \mbox{high-,} or \mbox{all-}~scales, the
    proposed complexity measure yields multi-indicators that can be used to construct
    partial orders if the relationships are too complex to be explained by linear order.

The proposed complexity measure attains {\sl absolute} zero for shapes in a non-trivial
    equivalence class including square.
This equivalence class, which emerges as a consequence of our constructions, makes
    compositionality explicit.
To exemplify, the measure attains absolute zero for certain square tiles obtained by
    translating the largest possible base square in the direction of two grid axes (\S
    \ref{sec:zero}).
For such square tiles, the size of the largest square that fits the shape is constant at
    every pixel.
That is, the proposed complexity measure respects information repetition.
Roughly speaking, a member of the zero-complexity class is maximally compressible and
    constructible.
Additionally, if the measure is used as a guide in an application, such as image
    segmentation, shape optimization or compression, preference may be steered towards the
    nearest member(s) of the zero-complexity equivalence class rather than a simple square. 

Recently, Fatemi {\sl et al.} \cite{Fatemi2016} studied recovering binary shape from its
    sampled representations such that the constructed image is constrained to regenerate
    the same representations.
A more recent work addressing shape from samples is Razavikia {\sl et al.}
    \cite{Razavikia2019}.
In this context, proposed shape complexity measure can be utilized as a guide for judging
    how difficult the reconstruction of a shape is.

A different but related problem of recent interest is quantifying the complexity of
    high-dimensional datasets for estimating their classification difficulty.
In that context, a growing number of works emphasize the role of the shape of the decision
    hyper-surface as a determinant of either how complex the data is \cite{HoBasu2002} or
    how robust its classification by a certain classifier~\cite{Fawzi2017}.
As another work in classification context, Varshney and Willsky~\cite{Varshney2010}
    measured their classifiers in terms of levelsets of the decision hypersurfaces
    geometrical complexity using ${\mathbf \epsilon}$-entropy.
It appears that measuring high-dimensional shapes is of importance in understanding
    classifiers and large scale data sets.
A particularly interesting claim by Fawzi {\sl et al.}~\cite{Fawzi2017} is that
    vulnerability to adversarial attacks is related to positive curvature of the decision
    boundary.
Fawzi {\sl et al.} further attributed the robustness of the popular deep networks to the
    flatness of the shape of the produced decision boundaries.

\section{Literature}
\noindent
Maragos~\cite{Maragos1989} proposed an entropy-like shape-size
    complexity measure by quantifying the deviation of the shape from a reference shape in
    a multi-scale way using binary morphological openings and closings with convex
    structuring elements of increasing sizes.
Nguyen and Hoang~\cite{Nguyen2014} exploit the properties of Radon transform to measure
    polygonality.
Niimi~{\sl et al.} \cite{Niimi1997} measured local complexity of bit plane images for data
    encryption purposes.
Gartus and Leder~\cite{Gartus2017} examined the relation between perceived symmetry of
    abstract binary patterns and computational measures.
Fan {\sl et al.}~\cite{Fan2017} examined visual complexity of binary ink paintings.
Zanette~\cite{Zanette2018} proposed a complexity measure for binary images based on
    diversity of the length scales in the depicted motifs.
The proposed measure penalizes images where gray levels are either distributed at random
    or ordered into a simple broad pattern.
Plotze {\sl et~al.}~\cite{Plotze2005}, for the purpose of plant species identification,
    defined complexity measures of leaf outline and veins. 
Complexity of binary images and shapes is of wide interest and examples can be enriched.

Page {\sl et al.}~\cite{icip2003Page} associated shape complexity via entropy of boundary
    curvature;
Chen and Sundaram~\cite{Chen2005} proposed correlates of Kolmogorov-complexity on boundary
    curvature, and used them to design efficient shape rejection algorithms
    \cite{Chen2005_rejection} that incorporate shape complexity; Chazelle and
    Incerpi~\cite{Chazelle84} related complexity for polygonal shapes to how entangled the
    polygon is.
Widespread use of boundary-curvature-related information as a correlate of complexity is
    due to Euclidean geometric consideration of complexity of shapes and seminal
    perception work \cite{Attneave,Arnoult1960} that established circle as the simplest.
Recently, Genctav and Tari, using curvature-dependent flows, proposed characterizing local
    circularity~\cite{Genctav2019} and ordered shapes based on entropy in curvature scale
    space~\cite{Genctav2017}.

The earliest interest in quantifying circularity can be traced back to the isoperimetric
    problem which is the problem of finding which contour, among those of equal
    perimeters, encloses the maximal area.
From this problem follows the form factor, a measure given by $4\pi A / P^2$ for a two
    dimensional shape of perimeter $P$ and area $A$, and is maximized for circles.
The form factor and its variants are reviewed by Ritter and Cooper~\cite{Ritter2009}. 
Rosin proposed several global measures of basic shapes~\cite{Rosin1999,Rosin2003}.

Measuring rectangularity is emerging as a specific interest due to
    applications.
Urban planning and landscape ecology is one area where rectangular shapes rather than
    circular ones form a reference~\cite{Moser2002landscape}.
Proper alignment of parts on a robotic production line is another application. 
In image segmentation applications, rectangularity measure can be used to improve
    over-segmented images~\cite{Ngo2015}.
The need for measuring high dimensional shapes is motivated by understanding large
    datasets and classifier behavior~\cite{{Varshney2010},Fawzi2017}.

\section{Method}
\label{sec:method}
\noindent 
Using concepts from mathematical morphology (structuring elements, erosions and distance
    transforms), we make the following observations:
\begin{itemize}
\item 
    For a shape $S$, successive upper levelsets of its Euclidean distance transform (EDT)
    correspond to
        its successive erosions using a disc structuring element.
\item 
    If we shrink the shape boundary $\partial S$ by moving its points at a speed
        proportional to the curvature and in the direction of inward normal, the boundary
        deforms towards a curve of uniform curvature, {\sl i.e.}, a circle.
    The regions closed by successive shape boundaries are adaptive erosions and
        form a multi-scale shape representation.
\item 
    If $\partial S$ is a circle, {\sl i.e}, $S$ has a uniform boundary curvature, its
        distance transform's upper levelsets agree with adaptive erosions.
    A $\partial S$ deviates from a circle, the discrepancy between the respective
        levelsets increase.
    The discrepancy is higher for those levelsets that are nearer to boundary and lower
        for those that are nearer to center.
\end{itemize}

\noindent
Suppose we are able to embed boundaries of successive adaptive erosions as levelsets of
    a field $f_S$ on $S$.
Then the congruency between $f_S$ and EDT can be measured by how
    uniform $f_S$ is at a levelset of of the latter.
If $S$ is a disc, the levelsets of the two agree. It can be said that the field implicitly
    encodes curvature \cite{Tari1996}.

The field construction can be extended to spaces other than Euclidean.
In that case, the field may encode some other concept, e.g. of having $90 \degree$ angles.

Intuitive idea is the following: Circle is the unit ball in Euclidean space.
Hence, in Euclidean space, circle is the simplest shape because one parameter (radius) is
    enough to construct it.
Likewise, under $\| . \|_\infty$ the unit ball is the square, which is the ideal shape
    that can be constructed as a binary image.
This motivates us to employ $L^\infty$ as the proper metric space to implicitly code
    relevant features.
 
The proposed method has two components: The first component is to compute $f_S$ in
    $L^\infty$.
The second component is to measure how well $f_S$ and the distance transform computed with
    the right metric agree.
In our case, it is $\| . \|_\infty$, also known as Chebyshev or chessboard distance.
In the rest of the paper, $t$ will be used to denote the normalized $\| . \|_\infty$
    distance transform for a shape.

%
%
%\begin{enumerate}
%\item 
%    constructing $f_S$ in $L^\infty$; 
%\item 
%    measuring how well $f_S$ and $t$ (computed with the right metric) agree. 
%\end{enumerate}
%
%   
    
\subsection{Constructing $f_S$}
\noindent 
Assuming the space in which shape $S$ is embedded has uniform grid, we solve the following
    PDE inside $S$
    \begin{equation}
        \left( \Delta_\infty - \frac{1}{\rho^2} \right) f_S
        = -1  
        \mbox{ subject to } 
        f_S \Big|_{\partial S} = 0 
    \label{eq:spe_inf}
    \end{equation}
    \noindent 
    where $\Delta_\infty$ is the Laplace operator in $L^\infty$. 
We note that $\Delta_\infty f$ is the minimizer of $\int |\nabla f|^p$ as $p\to \infty$.
Parameter $\rho$ is chosen to be an estimate of the shape radius as measured under $\| .
    \|_\infty$.
This choice ensures the robustness of solutions under changes in scale.
After its construction, $f_S$ is normalized to $[0,1]$, which renders the constant on the
    right hand side of \eqref{eq:spe_inf} devoid of meaning, so long as the sign is
    negative.
To acquire numerical solutions, the approximation to Laplace operator in
    $L^\infty$ \cite{Oberman2005},
    \begin{equation}
        \Delta_\infty f_S(x)
            \approx
        \max_{y\in B(x)} f_S(y) 
        +
        \min_{y\in B(x)} f_S(y)
            - 
        2 f_S(x)
        \label{eq:Lapinf}
    \end{equation}
    \noindent 
    where $B(x)$ denotes a unit ball centered at $x$, is used.
Note that the expression \eqref{eq:Lapinf} corresponds to difference between forward and
    backward morphological derivatives \cite{Maragos1996,BrockettMaragos1992}.

Points at a system governed by \eqref{eq:spe_inf} \textit{generates} and
    \textit{cumulates} the values of field.
Each point introduces an increment to the neighborhood average (which is
    $\left( \max f_S + \min f_S \right) $ in $L^\infty$).
Amount of increment is decided by the screening parameter $1/\rho^2$.
This increment creates levelsets increasing away from the boundary and these levelsets can
    be viewed as embeddings of gradual deformations of $\partial S$ towards a curve
    possessing features of the reference shape as determined by the ball of chosen metric
    space.
The maximum value of field is attained at the points with maximum distance to the
    boundaries, as they are the points of most cumulation.

\begin{figure}[h]
\centering
\begin{tabular}{cc}
    \includegraphics[width=.4\linewidth]
        {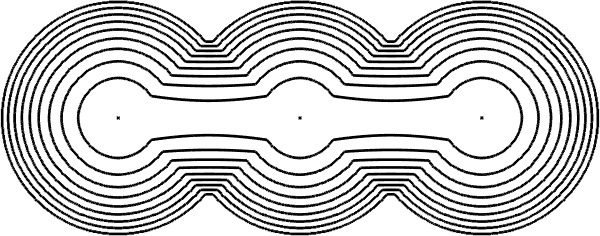} &
    \includegraphics[width=.4\linewidth]
        {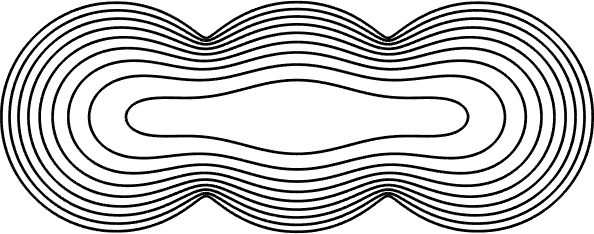} \\
        (a) & (b)
\end{tabular}
\caption{Local behavior of $L^\infty$ vs global behavior of $L^2$}
\label{fig:Linf_L2}
\end{figure}

An illustration of the levelsets of \eqref{eq:spe_inf} is given in
    Fig.~\ref{fig:Linf_L2}~(a).
Observe that the level curves become locally flat. For comparison, solution in $L^2$ is
    depicted in Fig.~\ref{fig:Linf_L2} (b).
In the later case, the level sets gets rounder.
%For solutions in $L^2$, this behavior is shown to be related to the
%    curvature of levelsets
%    \cite{Tari1996}. 

\begin{figure}[h]
    \centering
    \begin{tabular}{cccc}
        \includegraphics[height=.15\linewidth]
            {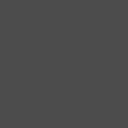} &
        \includegraphics[height=.15\linewidth]
            {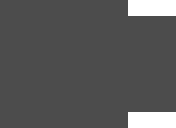} &
        \includegraphics[height=.15\linewidth]
            {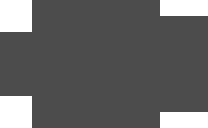} &
        \includegraphics[height=.15\linewidth]
            {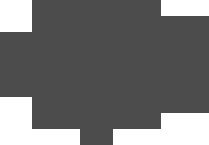} \\
        $S_0$ & $S_1$ & $S_2$ & $S_3$ \\[10px]
        \includegraphics[height=.15\linewidth]
            {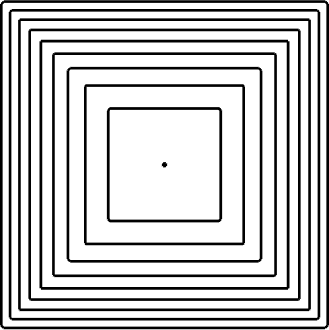} &
        \includegraphics[height=.15\linewidth]
            {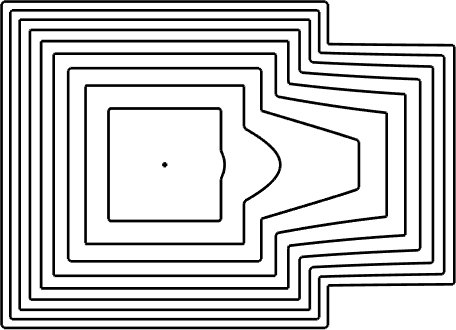} &
        \includegraphics[height=.15\linewidth]
            {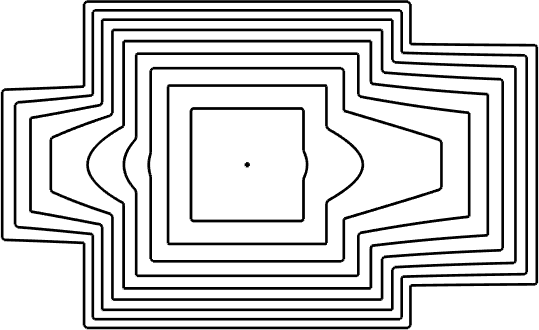} &
        \includegraphics[height=.15\linewidth]
            {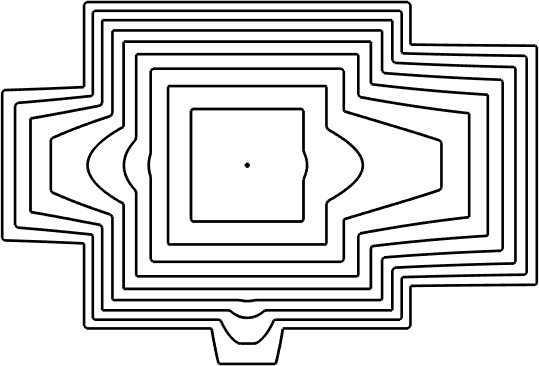} \\
        $f_{S_0}$ & $f_{S_1}$ & $f_{S_2}$ & $f_{S_3}$ \\
    \end{tabular}
    \caption{
        Successive addition of rectangular appendages (top row) and the levelsets of the
            acquired fields (bottom row)
    }
    \label{fig:appendages}
\end{figure}

Now let us consider (in $L^\infty$ setting) the effect of rectangular
    appendages on the field of a square (Fig.~\ref{fig:appendages}).
These appendages affect the field only in a region determined by their widths.
Outside these regions, they are disregarded.
The square $S_0$ has a side length of $128$ pixels.
The contacting width of the appendages are $96$, $64$, and $32$ pixels, respectively.

\subsection{Measuring How Well $f_S$ and $t$ Agree}
\label{ssec:congruence}
\noindent 
The field $f_S$ can be regarded as a \textit{well-behaving distance transform}. 
Recall that its levelsets agree with the levelsets of $t$ whenever the boundary 
    $\partial S$ is isotropic in the sense of chosen metric.
The discrepancy between $f_S$ and $t$ is due to the smoothed propagation of levelsets
    of $f_S$ in comparison to those of $t$.
To measure discrepancy, it is enough to measure deviation of the values of $f_S$ collected
    from a level set of $t$ from uniformity.
Thus, the problem becomes uniformity quantification.

\begin{figure}[h]
    \centering
    \includegraphics[width=.6\linewidth]
        {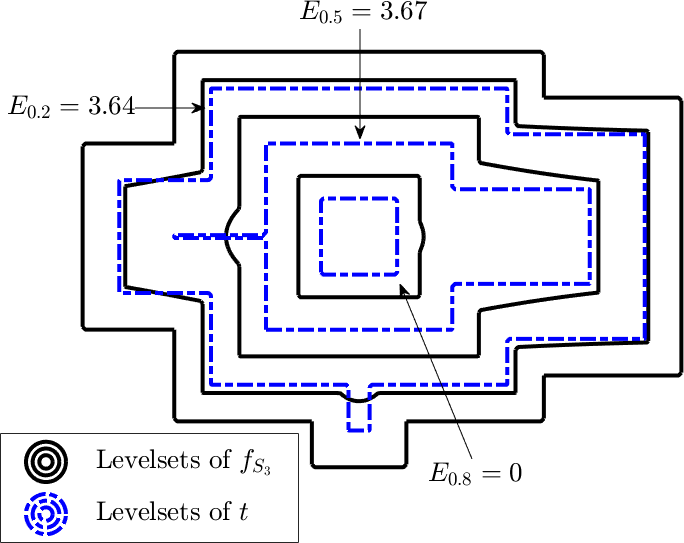}
    \caption{Levelsets of $f_S$ and $t$ for $S_3$}
    \label{fig:appendage_levelsets}
\end{figure}

The process is visualized in Fig.~\ref{fig:appendage_levelsets} using $S_3$ from
    Fig.~\ref{fig:appendages}. The black lines correspond to levelsets of $f_S $ at
    $0.01$, $0.3$, $0.6$, $0.9$, and blue dotted lines correspond to those of $t$ at
    $0.2$, $0.5$, $0.8$. $E_{t^\ast}$ is the uniformity estimated at $t=t^\ast$ using
    entropy.
To calculate entropy, we need to acquire a pseudo probability distribution.
For that purpose, the values of $f_S$ at a levelset of $t$ are partitioned into a fixed
    number of bins (in our implementation, $1024$), and then normalized.
        
\begin{figure}[h]
    \centering
    \begin{tabular}{c}
    \includegraphics[width=.6\linewidth]    {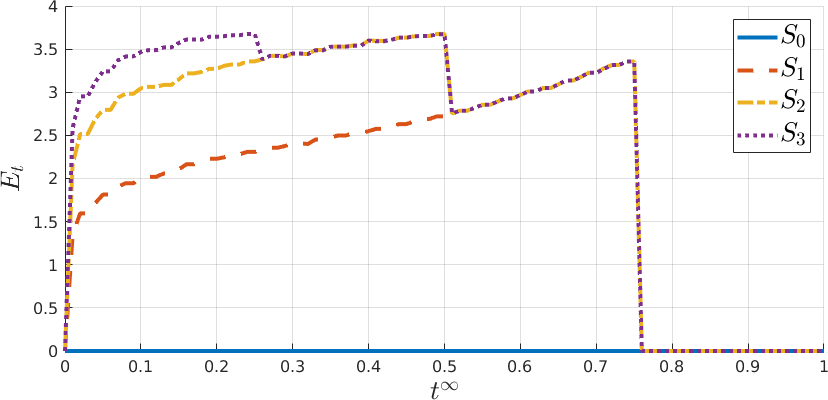}\\
     \includegraphics[width=.6\linewidth]   {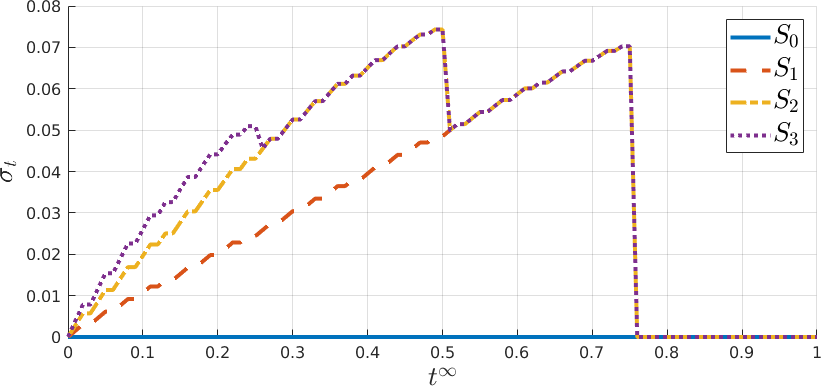} 
    \end{tabular}
    \caption{
        Scale versus complexity estimated using entropy (top) and standard deviation
            (bottom)
    }
    \label{fig:appendages_entropy_std}
\end{figure}

Our intuition is that any uniformity estimator could be used to obtain an ordinal number
    for complexity.
We experimented with both the entropy and standard deviation and observed that their
    behavior with respect to scale seemed to be equivalent even though their numerical
    values differ (Fig.~\ref{fig:appendages_entropy_std}).
In the rest of the paper, to avoid any confusion, we will mean the uniformity estimated
    using the entropy whenever we refer to shape complexity.

We can talk about a cutoff level $t_c$ of an appendage, such that for
    $t>t_c$ effect of the appendage on the field disappears.
$t_c$ is determined by the ratio of contacting width of the appendage to
    the width of the main body.
%\sout{
%    Value of $t_c$ is determined by the ratio of width of appendage to the main shape's
%    radius.
%}
For example, in Fig.~\ref{fig:appendages}, $t_c$ for the appendage introduced in $S_3$ is
    $32/128=0.25$, meaning that for $t>t_c=0.25$, $f_{S_2}$ and $f_{S_3}$ are identical.
Likewise, for $t>0.5$, we have $f_{S_1} = f_{S_2} = f_{S_3}$; and for $t>0.75$ we have
    $f_{S_0} = f_{S_1} = f_{S_2} = f_{S_3}$.
If we order the four shapes based on the uniformity of the levelsets, the order is
    {$S_0 < S_1 < S_2 < S_3$} for $t \in (0,    0.25]$; 
    {$S_0 < S_1 < S_2 = S_3$} for $t \in (0.25, 0.5 ]$;
    {$S_0 < S_1 = S_2 = S_3$} for $t \in (0.5,  0.75]$; 
    finally all shapes become equal after $t > 0.75$.
Though $t \in [0,1]$, it does not define a continuous scale parameter.
This is because $\| . \|_\infty$ distance transform produces integer values in
    $[0,\max]$.
Hence, $t$ admits only $\max$-many distinct scales.

\subsection{Implementation Details}
\noindent Discretizing \eqref{eq:spe_inf}, we obtain
\begin{equation}
    \max\limits_{y\in B(x)} f_S(y)
    +
    \min\limits_{y\in B(x)} f_S(y)
    - 
    \left(2+\frac{1}{\rho^2}\right) f_S(x)
    + 1
    = 
    0
    \label{eq:spe_discretized}
\end{equation}

\noindent 
The field $f_S$ constructed using the explicit scheme
\begin{equation}
    \begin{split}
        &\frac
        {
            f_S^{(k+1)}(x) - f_S^{(k)}(x)
        } {\Delta k}
        =\\[.1cm]
        &\quad
        {
            \max\limits_{y\in B(x)} f_S^{(k)}(y)
            +
            \min\limits_{y\in B(x)} f_S^{(k)}(y)
        }
        - \left(2+\frac{1}{\rho^2}\right) f_S^{(k)}(x)
        + 1
    \end{split}
    \label{eq:spe_iteration}
\end{equation}
\noindent
    where $f_S^{(k)}(x)$ is the constructed field at the $k$th step, and $f_S^{(0)}\equiv 0$.
Convergence conditions imposed on this scheme are
\begin{enumerate}
    \item maximum absolute value, $\max |\Delta f_S^{(k)}|$,
        of RHS of \eqref{eq:spe_iteration} is below some threshold, $\epsilon_1$,\vspace{3px}
    \item $\max |\Delta f_S^{(k+1)} - \Delta f_S^{(k)}| \le \epsilon_2$
\end{enumerate}
In our implementation $\epsilon_1=N \times 10^{-6}$ and $\epsilon_2=N \times 10^{-10}$ are
    used where $N$ is the number of nonzeros of the shape.
We can improve this solution by using our insight about the correlation of $f_S$ with the
    distance transform, and construct a systems based solution.
In doing so we use $t$ to guide us about the
    locations of local maxima and minima of the neighborhood of each point.

With this initial guess, we construct $A \vec{x} = b$ where $A$ is a sparse matrix
    and solve for $\vec{x}$.
Pointwise error of the acquired solution is calculated using the LHS of
    \eqref{eq:spe_discretized}, and is fedback to $\vec{x}$.
This process is iterated by using the $\vec{x}$ acquired in the last step as our next
    guess about the location of local maxima and minima.
Convergence condition for the iteration on the systems solution is either having
    \begin{enumerate}
        \item $\max \Delta f_S^{(k)}= 0$, or\vspace{3px}
        \item $\max |\Delta f_S^{(k+1)} - \Delta f_S^{(k)}| \le \epsilon_3$.
    \end{enumerate}
We used $\epsilon_3 = N \times 10^{-6}$ which is greater than $\epsilon_2$ since
    correction of small errors via systems solutions is more costly than correction via
    the purely iterative scheme.
If the iteration on systems solution converged with the second convergence condition,
    acquired solution is passed to the explicit scheme as the initial condition.

\section{Zero-Complexity Shapes}
\label{sec:zero}
\begin{figure}[h]
    \centering
    \includegraphics[width=.8\linewidth]   {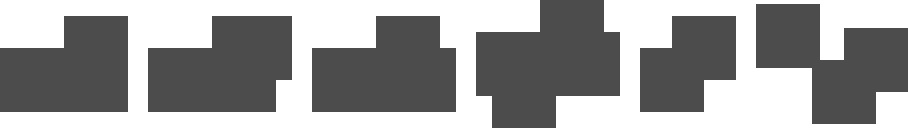}
    \caption{Sample zero-complexity shapes} 
    \label{fig:zerocomplexity} 
\end{figure}

\noindent 
Zero complexity implies that the level sets of $t$ and the constructed field $f_S$ are
    congruent.
Consider a shape obtained by adding a rectangular appendage to a base square.
If we imagine, at any pixel, comparing the value of $f_S$ to what the value would be if
    that pixel belonged to a square, the disagreement vanishes after the cutoff scale.
If the contacting width is equal to the side length of the square the formed shape becomes
    a rectangle, making the cutoff scale approach to one and the respective level sets
    congruent at any scale.
We can think of obtaining a rectangle from a base square as translating the center in the
    direction of the either of the grid axes.
Let $a, b \in \mathbb{Z}$ such that $a<b$ be the side lengths of our rectangle.
Consider adding to it a rectangular appendage of height $h$ with contacting width of $a$,
    making either a new rectangle with sides $a$ and $b+h$ or compound shape with two
    rectangles of sizes $a\times b$ and $a\times h$.
In either of the cases, congruence is unharmed.
These are the shapes with constant width and they can be constructed using identical
    square tiles without losing their qualities.
We can also translate the base square in the diagonal direction such that the two squares
    either overlap or touch.
The complexity is zero in either of the cases.
This because in the vicinity of the overlap or touch, the configuration is symmetric (in
    the sense of $L^\infty$ norm); consequently, level sets of $t$ and $f_S$ are congruent.
In the case of overlap, a neck joining the two squares is formed; hence, the resulting
    shape is not of constant width.
These shapes attain zero-complexity also in the shape-size complexity framework of Maragos
    \cite{Maragos1989}, if measured at the scale given by the size of the maximal square.
Illustrative examples are depicted in Fig.~\ref{fig:zerocomplexity}. 

A relevant claim by Donderi \cite{Donderi2006} is that the perceived complexity of a
    compound shape obtained by overlapping two shapes is the average of their
    dissimilarity and the average complexity value obtained by averaging perceived
    complexities of the individual shapes.
The behavior of our measure is consistent with this claim.
While earlier perception research links perceived complexity to features such as
    curvature, Donderi \cite{Donderi2006}, Forsythe {\sl et al.} \cite{Forsythe2011} and
    several others showed a correlation between compressed-file-size and perceived
    complexity.

\label{ssec:relative}
%%%%%%%%%%%%
\section{Experimental Results and Discussion}
\begin{figure}[h]
    \centering
    \includegraphics[width=.55\linewidth]   {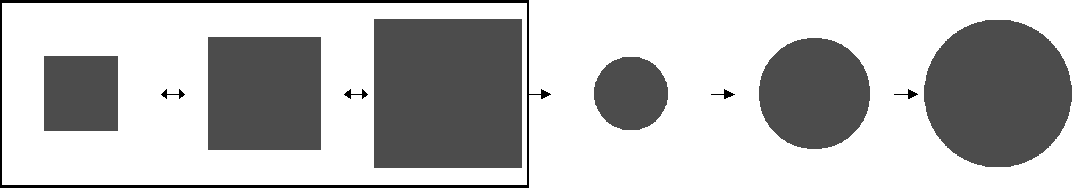}
    \caption{Boundary detail increases complexity} 
    \label{fig:compar_sqdi_1_LINFTY} 
\end{figure}

\noindent
In the first experiment, we compare squares and disks with varying sizes.
The order shown in Fig.~\ref{fig:compar_sqdi_1_LINFTY} is observed uniformly at all
    scales.
Single arrows are from simpler shapes to more complex ones and double arrows indicate
    equal complexity.
Whenever there is equivalence among a group of shapes, the equivalent shapes are
    displayed inside a rectangular box. 
Squares have the lowest (zero) complexity, regardless of their sizes, whereas disks are
    ordered by increasing size. 
This is the expected behavior for shapes in $\mathbb{Z}^n$.
The peak complexity value for the circles is close to $5$, and even for the smallest
    circle with $64$ pixel radius, $75\%$ of the time (scale), the value remains above
    $3$.
%Contrasting these complexity values with those of perceptually complicated bats from
%    MPEG7-dataset, or a real life rectangular floor plan reveals how complicated a circle
%    is as an object embedded in $\mathbb Z^n$ (Fig.~ \ref{fig:houseplan_complexity}).
In low radius limit, the digital circle becomes a cross hence a zero-complexity shape.

In the second experiment, we compare four rectangular shapes obtained by 
    adding four appendages of widths $32$
pixels to a base square with side length $128$ pixels. 
    \begin{figure}[h!]
    \centering
    \begin{tabular}{c}
    \includegraphics[width=.55\linewidth, clip]
        {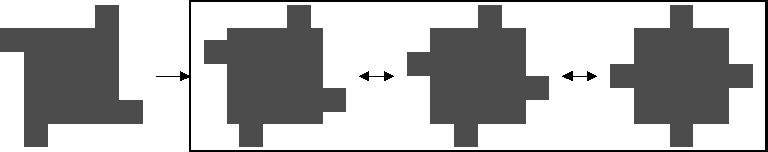}\\
%        (a)\\[3px]
%    \includegraphics[width=.9\linewidth, clip]
%        {figures/18_ds3_quad_Li_d=3_linear.png}\\
%        (b)
    \end{tabular}
    \caption{Varying the appendage position}
    \label{fig:compar_dataset_3_quad} 
\end{figure}
The ordering at lower scales, $t\in(0,0.25]$, is shown in
    Fig.~\ref{fig:compar_dataset_3_quad}.
As $t$ increases beyond the cutoff, $0.25 \,(\shorteq\,32/128)$, all four shapes attain
    the same complexity.
Any irregularity caused on the field occurs around joints because of the local behavior of
    $L^\infty$ solutions.
Since the shapes introducing sixteen corners (the last three shapes) cannot be
    distinguished around joints from one another locally, they have the same complexity.
The shape that introduces twelve vertices (the first shape from left), however, can be
    told apart since it has only four joints, in contrast to eight joints of the other
    three shapes.
 
\begin{figure}[h]
    \includegraphics[width=.97\linewidth]
        {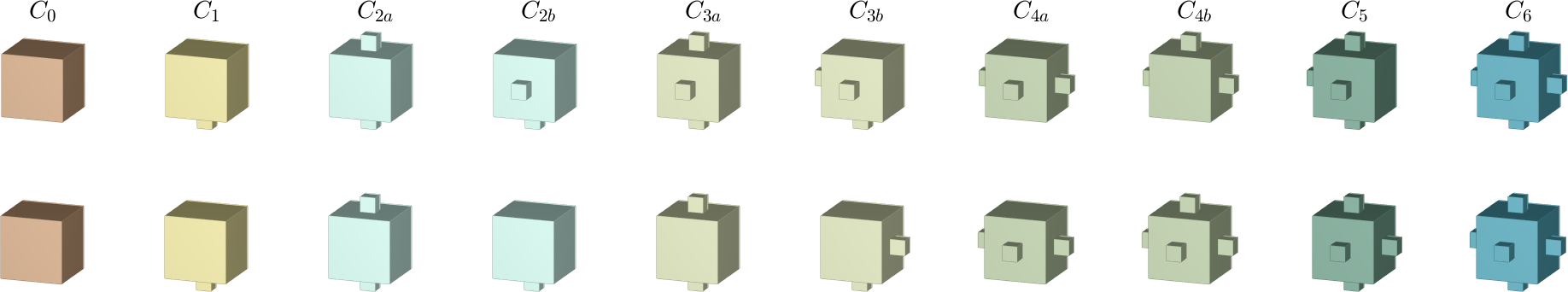}
    \caption{Cubes with appendages}
    \label{fig:cubes_64_sample}
\end{figure}

In the third experiment, we construct ten shapes by successively appending smaller cubes
    of side length $16$ voxel at the center of the surfaces of a larger cube of side
    length $64$ voxel. Five samples from the set are depicted in
    Fig.~\ref{fig:cubes_64_sample}.
The naming of the shapes indicate the number of appendages, and their locations, if
    needed.
Subscript $a$ indicates a preference towards opposite surfaces when appending a smaller
    cube, and $b$ indicates a preference towards adjacent surfaces.

\begin{figure}[h]
    \centering
    \includegraphics[width=.6\linewidth] {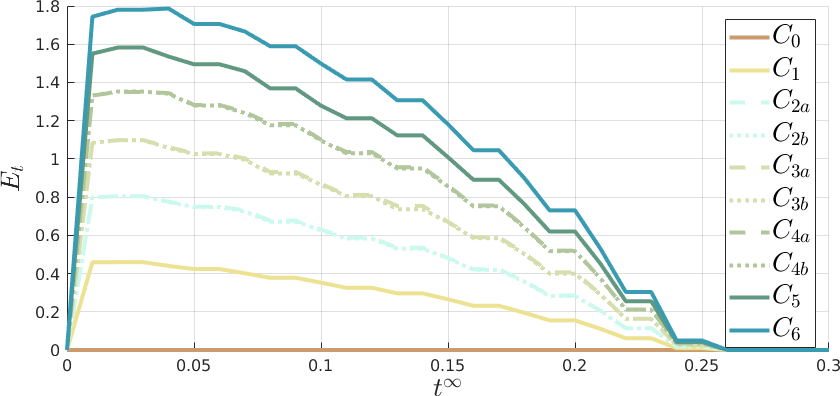}
    \caption{Scale versus complexity for cubes with appendages }
    \label{fig:cubes_64_entropy}
\end{figure}

Scale versus complexity plots in the interval $(0,0.3]$ are shown in
    Fig.~\ref{fig:cubes_64_entropy}.
$S_0$ is the simplest shape, as expected, with a complexity score of zero uniformly across
    all values of $t$ and the complexity increases with increasing amount of 
    appendages.
%For  $t^\infty \in \{0, 0.01, \dots, 0.3\}$. 
As the scale increases beyond the cutoff,  $0.25 \,(\shorteq\,16/64)$, all shapes attain
    zero complexity.
%
%The location of appendages has no observable effect on the measured complexities,
%     since their fields of interactions are not intersecting.
These results are in agreement with the results of the two-dimensional shapes.
%As a note on the implementation, using 26 immediate neighbors of a grid point as $B(x)$
%    in the discretization Eq.~\eqref{eq:Lapinf} is sufficient to construct the field at
%    three-dimensions.

\begin{figure}[ht]
  %  \centering
    \begin{tabular}{c}
    \includegraphics[width=.98\linewidth, clip]
        {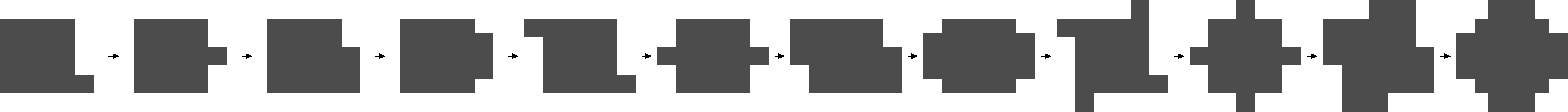}\\\\
   %     (a)\\[3px]
    \includegraphics[width=.98\linewidth, clip]
        {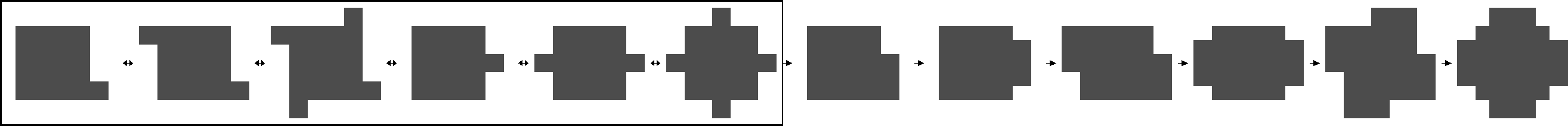}\\
%       (b)
       %\\[3px]
%    \includegraphics[width=.95\linewidth, clip]
%        {figures/21_ds4_union_Li_d=8_linear.png}\\
%        (c)
    \end{tabular}
    \caption{Varying the number, width and position}
    \label{fig:compar_dataset_4} 
\end{figure}

In the fourth experiment, we observe the joint effect of the placement, size and number of
    appendages.
We construct a set of shapes by varying these three properties. 
The appendage placement is chosen in two ways: at the center or the corner. 
The size (width) of appendages is also chosen in two ways: $32$ or $80$ pixels.
Finally, the number of the appendages is chosen in three ways: $1, 2$ or $4$.
Ordering obtained by our method using values collected at $t\in (0,1/4]$ is shown in the
    top row of Fig.~\ref{fig:compar_dataset_4}. 
The order is based on the number of appendages first, and between equals, width of
    appendages is taken into account.
That is, it induces a dictionary order.
Ordering obtained using values collected at $(1/4, 5/8]$ is shown in the bottom row of 
    Fig.~\ref{fig:compar_dataset_4}. 
All shapes with $32$ pixel width appendages attain the same complexity.
As $t$ increases further, all twelve shapes attain the same complexity.
%
%%%%
\subsection{Noisy Shapes} 
\label{ssec:noise}
\begin{figure}[h]
    \centering
    \begin{tabular}{c}
    \includegraphics[angle=90, width=.7\linewidth, trim={20px 0 170px 0}, clip]
        {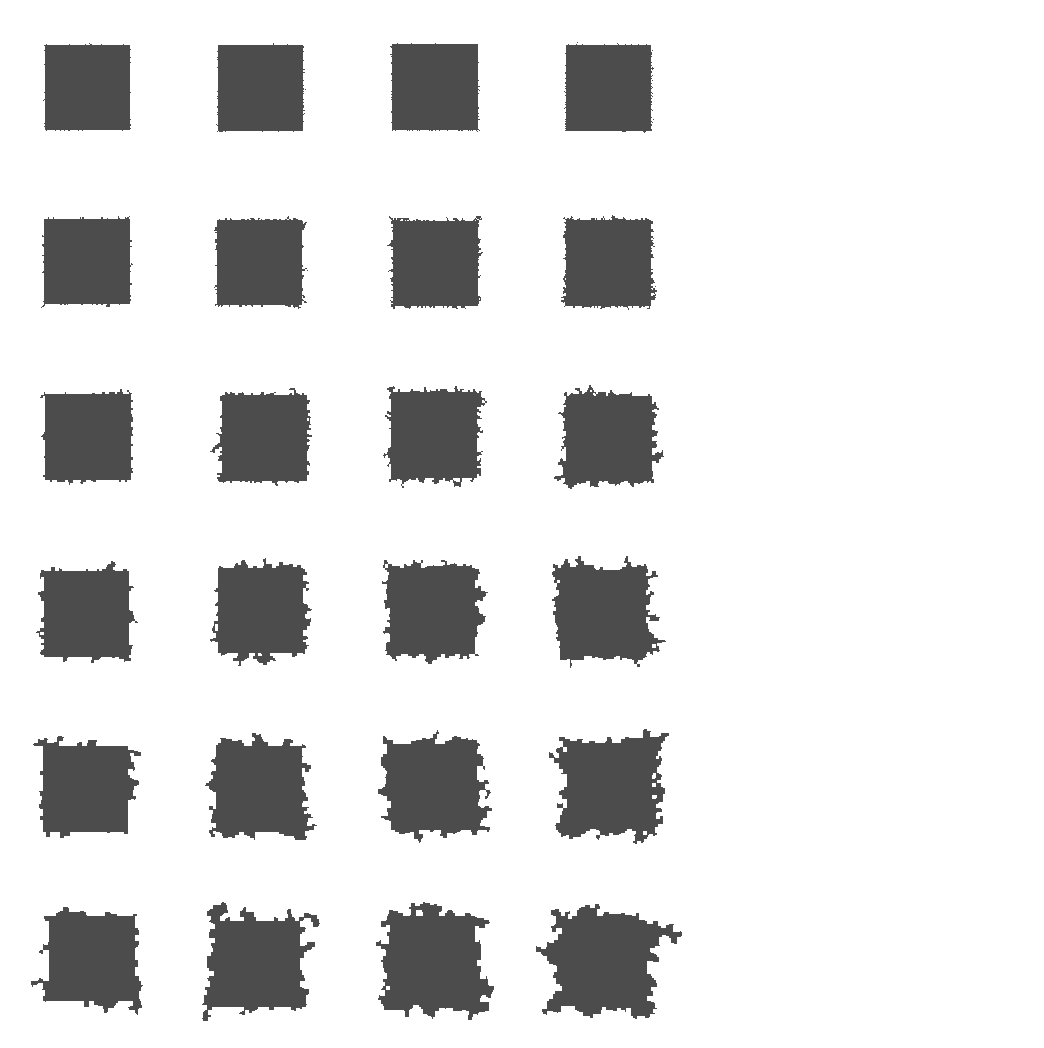} 
    \\[10px]
    \hline
    \\
    \includegraphics[angle=90, width=.7\linewidth, trim={15px 0 185px 0}, clip]
        {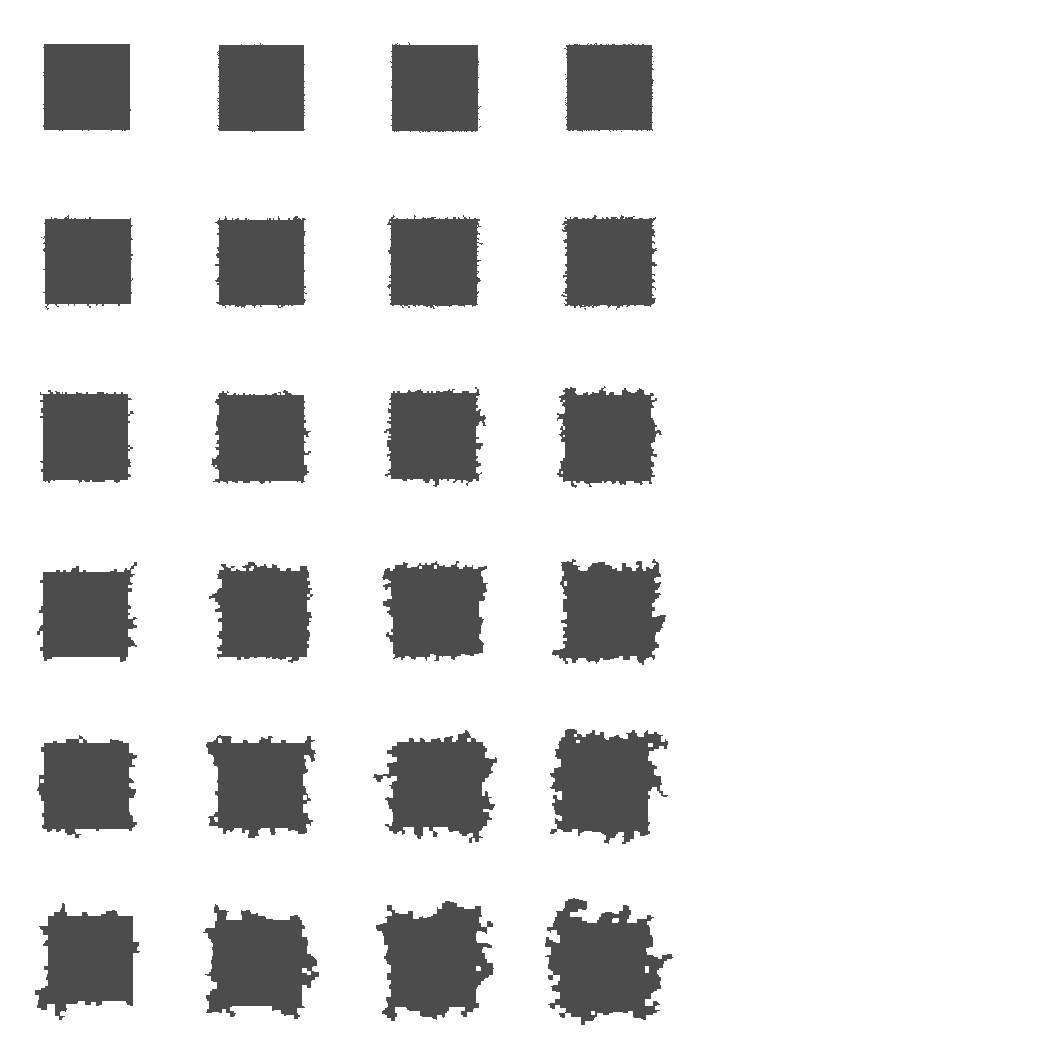}
    \end{tabular}
    \caption{
        Two random datasets. 
        Rows correspond to $\# = 400$ to $\# = 50$ from top to bottom.
        Columns correspond to $nf = 1$ to $nf = 6$ from left to right.
    }
    \label{fig:noise_datasets}
\end{figure}

\noindent
We created fifty random datasets with differing amount of noise
                    ($\# \in \{ 50,\, 100,\, 200,\, 400 \}$)
    and varying noise factors 
                    ($nf \in \{ 1,\, 2,\, 3,\, 4,\, 5,\, 6 \}$),
    which results in twenty-four shapes in each of the datasets.
The noise factor $nf$ determines the width and height of noise stochastically based on the
    radius of shape.
Pseudocode \verb|addNoise (.)| is given in Algorithm.~\ref{alg:addNoise}.
Two random datasets are shown in Fig.~\ref{fig:noise_datasets}. 
\begin{algorithm}
    \caption{Noise generation}
    \begin{lrbox}{\strel}
    \verb!strel(`square', nf)!
    \end{lrbox}
	\begin{algorithmic}
        \Function{addNoise}{$S$, $nf$} 
        \Comment{adds noise to shape $S$ with noise factor $nf$}
        \State $width \gets$ \Call {normRand}{$nf$, $nf/3$}
        \Comment{$\mu= nf$, $\sigma = nf/3$}
        \State $x_n \gets$ \Call {normRand}{$nf$, $nf / 3$, $width$}
        \Comment{returns $round(width)$ many random numbers}
        \State $y_n \gets$ \Call {normRand}{$nf$, $nf / 3$, $width$}
        \State $P \gets$ \Call {pointOnBoundary} {$S$}
        \Comment{returns the boundary pixels of shape}
        \State $S' \gets$ \Call {drawLinesX} {$S, P, x_n$}
        \Comment{puts $i\in x_n$ pixels along $x$-direction to the outside of $S$ around a pixel of $P$}
        \State $S' \gets$ \Call {drawLinesY} {$S',P, y_n$}
        \Comment{puts $i\in y_n$ pixels along $y$-direction}
        \State $S' \gets$ \Call {morphologicalClosing}{$S', square_{nf}$}
        \Comment{$square_{nf} = $ \usebox{\strel} }
		\State \Return $S'$
		\EndFunction \\
	\end{algorithmic}
    \label{alg:addNoise}
\end{algorithm}

In the first experiment, we fix the noise factor and then at each noise factor calculate
    modified Kendall~$\tau$ correlation between the expected order (\textit{i.e.} the
    order with respect to number of noise applications) and the obtained order.
Correlations are calculated for all fifty random datasets.
The top four rows of Table~\ref{table:noise_scores} show the results for each noise
    factor.
The entries are averaged modified Kendall~$\tau$ scores.
Respective standard errors of the estimation are depicted in the top four rows of
    Table~\ref{table:noise_stds}.
%\sout{acquired from noisy datasets using a measurement based on Kendall $\tau$.}
%
We note that the modified Kendall~$\tau$ returns $+1$ for equal pairs regardless of the
    reference ordering.
This is because in this dataset equal complexity values are observed to arise only when
    the complexity is zero, which means that added noise is completely disregarded.

\begin{table}[h]
    \centering
    \begin{tabular}{|c|*{9}{m{.5cm}}|}
    \hline
\diagbox[width=1.2cm]{$nf$}{$t$}
        &   $0.1$   &   $0.2$   &   $0.3$   &   $0.4$   &   $0.5$   &   $0.6$   &   $0.7$   &   $0.8$   &   $0.9$   \\
    \hline
    $1$ &   $1.00$  &   $1.00$  &   $1.00$  &   $1.00$  &   $1.00$  &   $1.00$  &   $1.00$  &   $1.00$  &   $1.00$  \\
    $2$ &   $0.82$  &   $0.95$  &   $1.00$  &   $1.00$  &   $1.00$  &   $1.00$  &   $1.00$  &   $1.00$  &   $1.00$  \\
    $3$ &   $0.96$  &   $0.94$  &   $0.92$  &   $0.94$  &   $1.00$  &   $1.00$  &   $1.00$  &   $1.00$  &   $1.00$  \\
    $4$ &   $1.00$  &   $0.95$  &   $0.87$  &   $0.85$  &   $0.88$  &   $0.93$  &   $0.98$  &   $1.00$  &   $1.00$  \\
    $5$ &   $0.99$  &   $0.95$  &   $0.91$  &   $0.87$  &   $0.83$  &   $0.87$  &   $0.86$  &   $0.94$  &   $0.98$  \\
    $6$ &   $0.97$  &   $0.96$  &   $0.91$  &   $0.87$  &   $0.83$  &   $0.81$  &   $0.83$  &   $0.88$  &   $0.93$  \\
    \hline
\diagbox[width=1.2cm]{$\#$}{$t$}
        &   $0.1$   &   $0.2$   &   $0.3$   &   $0.4$   &   $0.5$   &   $0.6$   &   $0.7$   &   $0.8$   &   $0.9$   \\
    \hline
   $50$ &   $0.99$  &   $0.96$  &   $0.97$  &   $0.99$  &   $0.99$  &   $1.00$  &   $0.99$  &   $1.00$  &   $1.00$  \\
  $100$ &   $0.99$  &   $0.98$  &   $0.95$  &   $0.95$  &   $0.96$  &   $0.97$  &   $0.98$  &   $1.00$  &   $1.00$  \\
  $150$ &   $1.00$  &   $0.98$  &   $0.96$  &   $0.96$  &   $0.96$  &   $0.95$  &   $0.97$  &   $0.98$  &   $0.99$  \\
  $200$ &   $0.99$  &   $0.99$  &   $0.97$  &   $0.95$  &   $0.94$  &   $0.94$  &   $0.95$  &   $0.97$  &   $0.98$  \\
    \hline
\end{tabular}
\vspace{2px}
\caption{Kendall~$\tau$ scores averaged over the fifty datasets}
\label{table:noise_scores}
\end{table}

\begin{table}[h]
    \centering
    \begin{tabular}{|c|*{9}{m{.5cm}}|}
    \hline                                                                                           
\diagbox[width=1.2cm]{$nf$}{$t$}
        &   $0.1$   &   $0.2$   &   $0.3$   &   $0.4$   &   $0.5$   &   $0.6$   &   $0.7$   &   $0.8$   &   $0.9$   \\
    \hline
    $1$ &   $0.00$  &   $0.00$  &   $0.00$  &   $0.00$  &   $0.00$  &   $0.00$  &   $0.00$  &   $0.00$  &   $0.00$  \\
    $2$ &   $0.22$  &   $0.12$  &   $0.00$  &   $0.00$  &   $0.00$  &   $0.00$  &   $0.00$  &   $0.00$  &   $0.00$  \\
    $3$ &   $0.11$  &   $0.13$  &   $0.19$  &   $0.15$  &   $0.00$  &   $0.00$  &   $0.00$  &   $0.00$  &   $0.00$  \\
    $4$ &   $0.00$  &   $0.12$  &   $0.21$  &   $0.28$  &   $0.24$  &   $0.16$  &   $0.10$  &   $0.00$  &   $0.00$  \\
    $5$ &   $0.07$  &   $0.12$  &   $0.15$  &   $0.18$  &   $0.23$  &   $0.19$  &   $0.28$  &   $0.13$  &   $0.08$  \\
    $6$ &   $0.09$  &   $0.11$  &   $0.15$  &   $0.20$  &   $0.24$  &   $0.24$  &   $0.25$  &   $0.19$  &   $0.14$  \\
    \hline
\diagbox[width=1.2cm]{$\#$}{$t$}
        &   $0.1$   &   $0.2$   &   $0.3$   &   $0.4$   &   $0.5$   &   $0.6$   &   $0.7$   &   $0.8$   &   $0.9$   \\
    \hline
   $50$ &   $0.03$  &   $0.06$  &   $0.06$  &   $0.05$  &   $0.05$  &   $0.00$  &   $0.03$  &   $0.00$  &   $0.00$  \\
  $100$ &   $0.03$  &   $0.05$  &   $0.08$  &   $0.07$  &   $0.07$  &   $0.07$  &   $0.05$  &   $0.00$  &   $0.00$  \\
  $150$ &   $0.00$  &   $0.05$  &   $0.08$  &   $0.07$  &   $0.07$  &   $0.09$  &   $0.07$  &   $0.05$  &   $0.03$  \\
  $200$ &   $0.03$  &   $0.03$  &   $0.06$  &   $0.07$  &   $0.07$  &   $0.09$  &   $0.08$  &   $0.06$  &   $0.05$  \\
    \hline                                                                                           
\end{tabular}
\vspace{2px}
\caption{Standard deviation of the Kendall~$\tau$ scores of fifty datasets}
\label{table:noise_stds}
\end{table}

In the next experiment, we fix the number of noise addition and then at each number
    calculate modified Kendall~$\tau$ correlation between the expected order
    (\textit{i.e.} the order with respect to noise factor) and the obtained one.
The results are given in the respective bottom four rows of
    Table~\ref{table:noise_scores} and Table~\ref{table:noise_stds}.
%In contrast to the number of noise additions which is a deterministic variable, the noise
%    factor is a random variable.
%Hence, the orderings based on $nf$ do not necessarily suggest a complexity based
%    order.
%This is reflected by the standard deviations of relevant measurements: Kendall~$\tau$
    %scores are lower than those acquired for fixed $nf$.

The drop in the scores with increasing $nf\,$ and $\#$ can be explained through the
    observation that the shapes start to lose their squareness with excessive
    noise.
The agreement between the induced and expected orderings increase with
    $t$, which is in accordance with the claim that effects of noise are disregarded
    at higher $t$s.

\subsection{Line Drawings of Floor Plans}
\label{ssec:drawings}
\begin{figure}[h]
    \centering
    \begin{tabular}{ccccc}
        \includegraphics[width=.13\linewidth]   {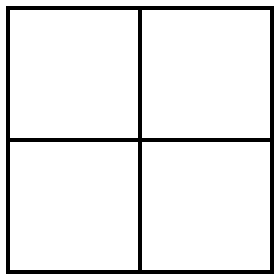} &
        \includegraphics[width=.13\linewidth]   {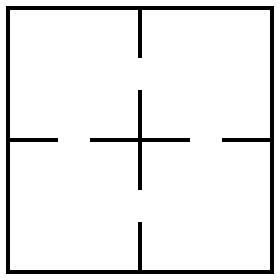} &
        \includegraphics[width=.13\linewidth]   {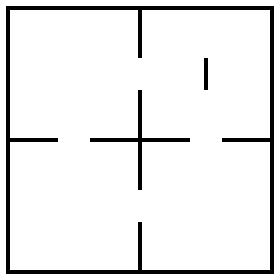} &
        \includegraphics[width=.13\linewidth]   {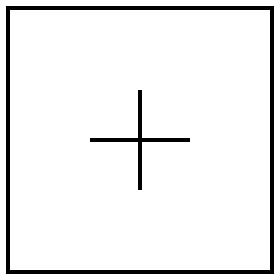} &
        \includegraphics[width=.2\linewidth]{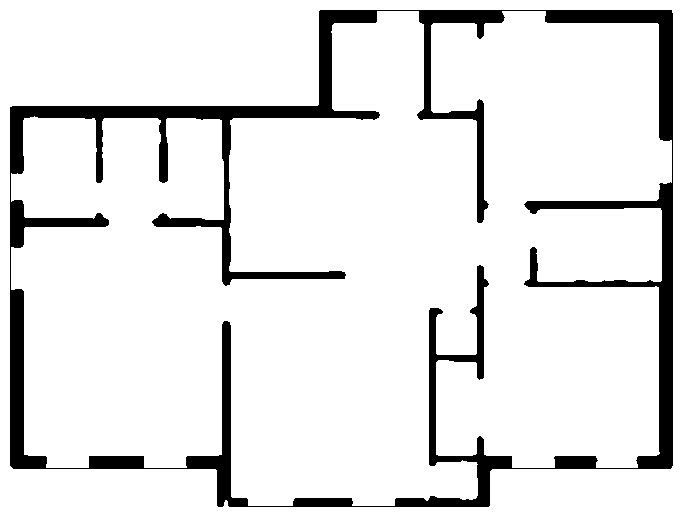} \\
        $P_0$ & $P_1$ & $P_2$ & $P_3$ & $P_4$
    \end{tabular}
    \caption{House plans}
    \label{fig:houseplans}
\end{figure}

\noindent 
We started with a rectangular drawing of a floor plan and then constructed four simpler
    cases. 
All five drawings are displayed in Fig.~\ref{fig:houseplans}.
We inserted them in a frame where lines (shown in black) serve as artificial shape
    boundaries.
The fields $f_S$ and $t$ are constructed in the white regions.
%As simpler references, we construct $P_0$, $P_1$, $P_2$ and $P_3$. 
%Boundaries of these shapes are displayed in Fig.~\ref{fig:houseplans}.
$P_0$ is a plan of four disconnected identical rooms of side lengths $128$ pixels.
$P_1$ is constructed by connecting the rooms with apertures of $32$ pixels.
$P_2$ is constructed by adding an obstacle to one of the rooms aligned with the vertical
    aperture and of size $32\times 4$ pixels.
$P_3$ is constructed by expanding the length of apertures of $P_2$ to $80$ pixel.

\begin{figure}[h]
    \centering
    \includegraphics[width=.6\linewidth] {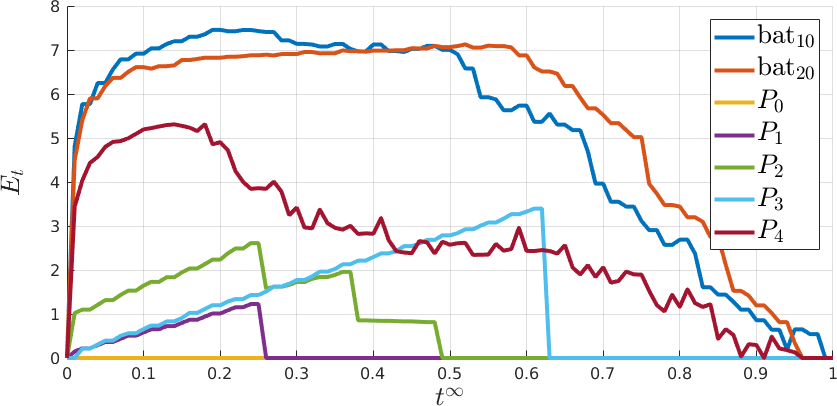}
    \caption{Scale versus complexity for floor plans}
    \label{fig:houseplan_complexity}
\end{figure}

In Fig.~\ref{fig:houseplan_complexity} scale versus complexity plot for the five line
    drawings are shown.
For comparison, we also include two complex shapes, $bat_{10}$ and $bat_{20}$ (bat
    silhouettes taken from MPEG7-dataset).
The complexity of $P_4$ is higher than that of $P_0$, $P_1$, and $P_2$ at all scales, and
    is lower than that of $bat_{10}$ and $bat_{20}$ at all scales except at $t=0.95$ where
    the complexity of $P_4$ is higher than $bat_{10}$'s.
The complexity of $P_3$ is higher than $P_4$'s at $t \in
    \{0.43,\,0.44,\,0.45,\,0.47,\,0.48,\,\dots,0.62\}$ and drops to 0 at $t = 0.63$.
This drop is explained by the cutoff level $t_c = 0.625 \,(\shorteq\,80/128)$ of the
    apertures of $P_3$.

%
%%%%%%

\subsection{Complex Relations Between Shapes: Incomparable Pairs}
\label{ssec:porder}
\noindent 
As the relation between a pair of shapes gets more complex than adding a rectangular
    appendage, so does their order relation which no longer exhibits a monotonic behavior
    with respect to scale.
For example, let us compare the complexity of plots of $P_2$ and $P_3$ using
    Fig.~\ref{fig:houseplan_complexity}.
%and $bat_{10}$ versus $bat_{20}$. For example, 
Integrating over all scales in order to obtain single number representing complexity, we
    can claim that $P_3$ is more complex than $P_2$.
But at lower scales where the boundary features dominate, $P_3$ is simpler than $P_2$.
It is only after a certain scale, when the effect of the obstacle in the upper right room
    is disregarded, that the order is reversed.
There are also other incomparable pairs: $P_3$ and $P_4$ or $bat_{10}$ and $bat_{20}$.
Indeed, it is impossible to linearly order all shapes.

It is, however, of important practical concern to compare shapes based on different
    complexity considerations.
Under such circumstances, although a linear order can not be established on the set of all
    shapes, it is possible to establish a partial order.
Partial order mainly differs from linear order by the presence of incomparable pairs.
From a given partial order, subsets on which a linear order is possible -{\sl chains}- can
    be extracted.
For example, using the measure integrated over low scales, $t \in \left(0, 0.25 \right]$,
    together with the measure integrated over all scales, yields partial order where $P_2$
    and $P_3$ as well as $bat_{10}$ and $bat_{20}$ are incomparable. $P_0$ is the least
    complex shape.
%Four chains each having five elements (e.g. $P_0 , P_1, P_3 , P_4, bat_{10}$) can be extracted. 
  
%  \begin{itemize}
%  \item $P_0 < P_1 < P_2 < P_4 < bat_{10}$
%   \item $P_0 < P_1 < P_3 < P_4 < bat_{10}$
%\item $P_0 < P_1 < P_2< P_4 < bat_{20}$
%  \item $P_0 < P_1 < P_3 < P_4 < bat_{20}$
%  \end{itemize}
  
\begin{figure}[hbt]
    \centering
    {
        \transparent{.8} 
        \includegraphics[width=\linewidth] {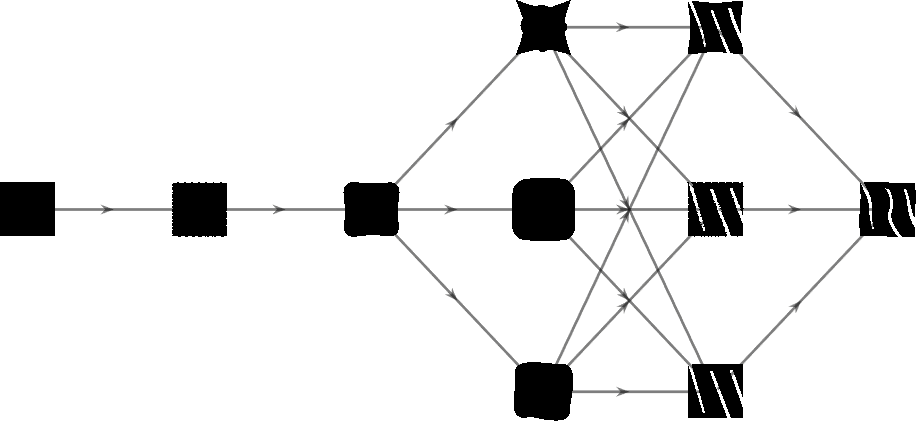}
    }
    \caption{Partial order graph using low-scale and all-scale}
    \label{fig:porder}
\end{figure}

As another example where partial order is more natural than linear order, we considered a
    set of $10$ shapes from device3 category of MPEG7-dataset. 
Using the same two indicators for measuring complexity (low-scale and all-scale), we
    constructed a partial order whose graph (Hasse diagram) is depicted in Fig.
    \ref{fig:porder}. 
The first three shapes are ordered linearly based on the deformations of square.
The third shape is followed by three others in different branches, each with a
    different kind of deformation: more concave, more oval, or more curved.
All of them are considered less complex than the shapes of the cut group.
Among the shapes of the cut group, the most complex is the one with curvy cuts. 
%Nine chains, each having six elements, can be constructed.

\section{Summary and Concluding Remarks}
\noindent 
Measuring complexity of shapes is important for a variety of reasons.
Though the problem is an old one, it lacks robust rigorous solutions because the concept
    is ill-defined, and biases due to Euclidean geometry are not necessarily useful in
    digital setting.

We set forth a clear goal in terms of constructibility in digital setting.
This motivated us to consider squares (hypercubes in high dimensions) rather than circles
    as the simplest. 
Using the link between $L^\infty$ metric and squares (cubes and hypercubes in higher
    dimensions) we constructed a field, $f_S$, whose levelsets encode squareness-adapted
    multi-scale simplifications.
We defined a multi-scale complexity measure by estimating the uniformity of $f_S$
    restricted to a certain levelset of $t$.
We observed an emergent cutoff scale above which the effect of a boundary detail vanishes.
This resulted in different shapes to converge in complexity at higher scales.
Furthermore, integrating the complexity over intervals of the scale parameter, we obtained
    multiple indicators that can be used to construct partial orders. 

An interesting future work is to identify local shape elements that increase complexity
    and suggest a complexity reducing decomposition.

\bibliographystyle{unsrt}

%\begin{thebibliography}{1}
%
%	\bibitem{kour2014real}
%	George Kour and Raid Saabne.
%	\newblock Real-time segmentation of on-line handwritten arabic script.
%	\newblock In {\em Frontiers in Handwriting Recognition (ICFHR), 2014 14th
%			International Conference on}, pages 417--422. IEEE, 2014.
%
%	\bibitem{kour2014fast}
%	George Kour and Raid Saabne.
%	\newblock Fast classification of handwritten on-line arabic characters.
%	\newblock In {\em Soft Computing and Pattern Recognition (SoCPaR), 2014 6th
%			International Conference of}, pages 312--318. IEEE, 2014.
%
%	\bibitem{hadash2018estimate}
%	Guy Hadash, Einat Kermany, Boaz Carmeli, Ofer Lavi, George Kour, and Alon
%	Jacovi.
%	\newblock Estimate and replace: A novel approach to integrating deep neural
%	networks with existing applications.
%	\newblock {\em arXiv preprint arXiv:1804.09028}, 2018.
%
%\end{thebibliography}

\end{document}